\documentclass[10pt,conference]{IEEEtran}
\usepackage{pifont}
\newcommand{\xmark}{\ding{55}}
\usepackage[utf8]{inputenc}
\usepackage{float}
\usepackage{hyperref}

\usepackage{amsmath,amssymb,amsfonts}
\usepackage{algorithmic}
\usepackage{graphicx}
\usepackage{textcomp}
\usepackage[table]{xcolor}
\usepackage[numbers]{natbib}
\usepackage[nameinlink,capitalise,noabbrev]{cleveref}
\usepackage{booktabs}
\usepackage{algorithm}
\usepackage{makecell}
\usepackage{wrapfig}
\usepackage{multirow}
\begin{document}

\title{TPSO: Training-Free Diverse Image Generation via Semantic Prompt Embedding Optimization}

\author{
\IEEEauthorblockN{
Debin Meng$^{1}$, 
Chen Jin$^{2}$, 
Zheng Gao$^{1}$,
Yanran Li$^{3}$,
Ioannis Patras$^{1}$, 
Georgios Tzimiropoulos$^{1}$
}
\IEEEauthorblockA{
\small $^1$Queen Mary University of London, $^2$Centre for AI, AstraZeneca, UK, $^3$University of Bedfordshire\\
{\tt\small \{debin.meng, z.gao, i.patras, g.tzimiropoulos\}@qmul.ac.uk}, \tt\small chen.jin@astrazeneca.com
\\ \tt\small Yanran.Li@beds.ac.uk
}
}

\maketitle

\begingroup
\renewcommand{\thefootnote}{}
\footnotetext{\textcopyright~2026 IEEE.}
\endgroup

\begin{abstract}
Image diversity remains a fundamental challenge for text-to-image diffusion models.
Low-diversity generation often leads to repetitive outputs, increasing sampling redundancy and hindering both creative exploration and downstream applications.
A key factor is the tendency of diffusion models to collapse toward \emph{strong modes in the learned distribution}.
Existing attempts to improve diversity, such as steering-based guidance, often introduce distortions that degrade image quality.
To address this issue, we propose \textbf{Token-Prompt Embedding Space Optimization (TPSO)}, a training-free and model-agnostic module.
TPSO introduces learnable parameters to explore underrepresented regions of the token embedding space, reducing the tendency to repeatedly sample from strong modes of the distribution.
Meanwhile, a prompt-level semantic constraint regulates distribution shifts, preventing quality degradation while preserving semantic fidelity.
Extensive experiments on MS-COCO across three representative diffusion backbones demonstrate that TPSO substantially improves diversity, boosting performance from 1.10 to 4.18, while maintaining image quality with only a modest inference-time overhead of 3.6\% to 8.9\%. \textit{Code is available at: \url{https://github.com/Open-Debin/TPSO}~.}
\end{abstract}

\begin{IEEEkeywords}
Text-to-Image Generation, Generative Diversity, Diffusion Models, Prompt Embedding Optimization, Training-Free Methods.
\end{IEEEkeywords}

\begin{figure*}[t]
    \centering
    \includegraphics[width=0.9\textwidth]{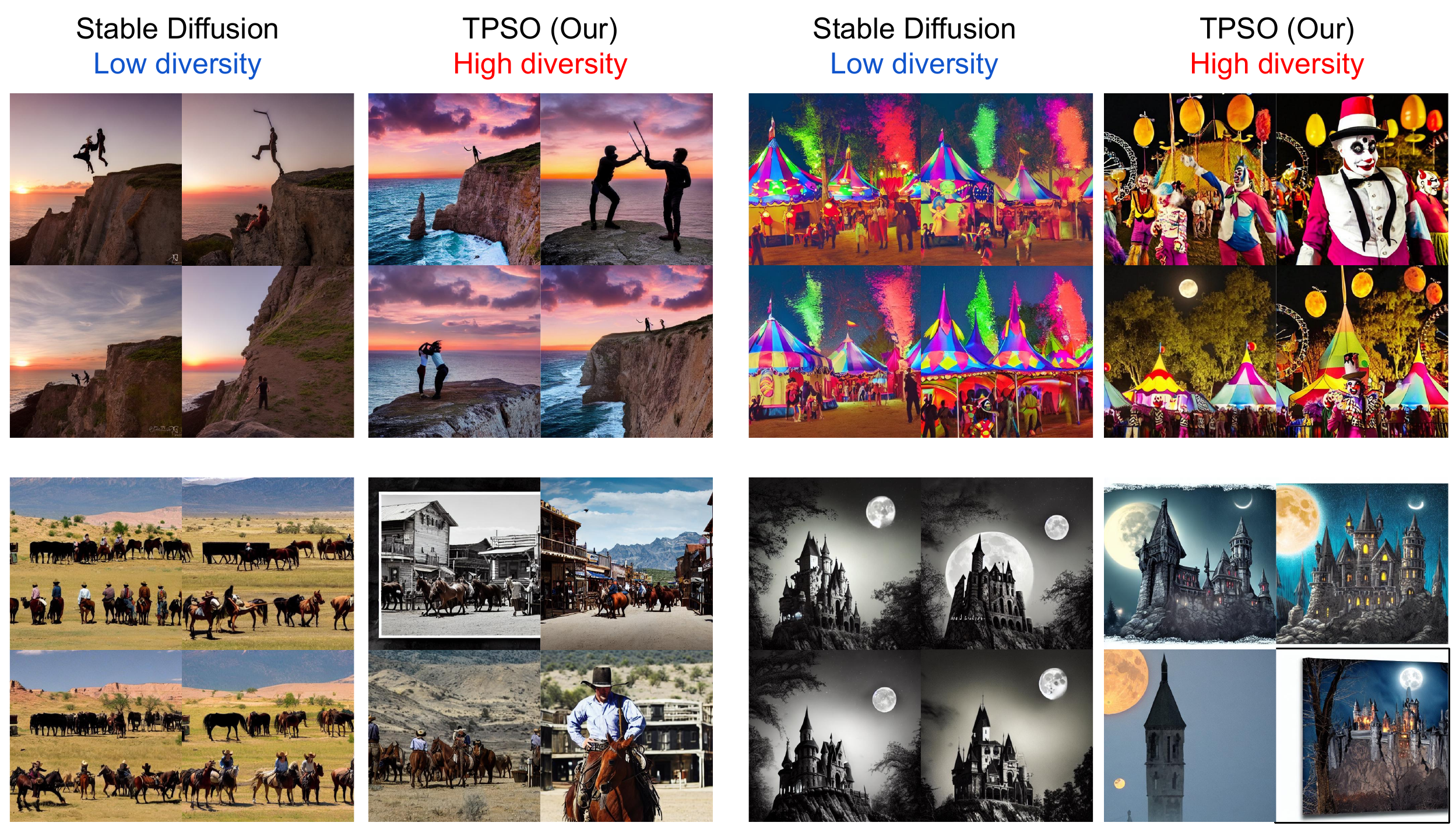}
    \caption{
        Baseline diffusion models tend to produce low-diversity or nearly repeated outputs across sampling runs under the same condition, 
        leading to redundant sampling efforts. 
        In contrast, our approach enhances output diversity, yielding a broader range of plausible and distinct visual outcomes.
    }
    \label{fig:teaser_compare}
\end{figure*}

\section{Introduction}
\label{sec:intro}
Powered by recent developments in diffusion models~\citep{ho2020denoising, song2020denoising} and contrastively trained text encoders~\citep{Radford2021LearningTV}, modern text-to-image systems are now capable of producing visually compelling and semantically faithful images from diverse natural-language prompts~\citep{rombach2022high}. A fundamental attribute of generative models is their ability to produce diverse outputs under the same condition. Ideally, a generative model should generate samples that are both consistent with the conditioning input and diverse in appearance. For example, a prompt describing “a wooden chair” can produce many valid outcomes that differ in shape, material texture, or stylistic design. However, Astolfi et al.~\cite{astolfi2024consistency} point out that current diffusion models often suffer from limited diversity: state-of-the-art Text-to-Image models tend to improve consistency and/or realism by sacrificing representation diversity. This lack of diversity leads to highly similar outputs under the same conditions, resulting in redundant resampling and increased computational cost, while also constraining creative exploration and impairing downstream applications such as data generation for model training, where repetitive samples reduce dataset utility. CADS~\cite{sadat2023cads} and Particle Guidance~\cite{corso2023particle} attribute this phenomenon largely to mode collapse induced by classifier-free guidance (CFG)~\cite{ho2022classifier}, a standard practice in diffusion models for enforcing fidelity. Under the strong guidance of CFG~\cite{ho2022classifier} tends to push the mini-batch samples towards a strong mode within the learned distribution to enhance fidelity, at the expense of diversity~\cite{corso2023particle}.

To address the limited diversity caused by CFG~\cite{ho2022classifier}, Particle Guidance~\cite{corso2023particle} optimizes intermediate latents within a batch by repelling samples from each other to enforce separation. Sadat et al.~\cite{sadat2023cads} further point out that the limited diversity of CFG stems from strong conditional guidance, which pushes samples toward strong modes of the learned distribution. Based on this insight, they propose Dynamic Classifier-Free Guidance (CFG) and CADS. Dynamic CFG~\cite{sadat2023cads} reduces the effect of CFG by dynamically modifying the guidance strength, while CADS~\cite{sadat2023cads} adjusts the conditional strength by interpolating Gaussian noise with the conditional signal. Although these approaches can improve diversity, they lack semantic constraints and therefore often push samples toward semantically implausible regions, which frequently introduces unintended distortions and degrades image quality (See~Fig. ~\ref{fig:quality_decay}).

To address these limitations, we introduce Token–Prompt Embedding Space Optimization (TPSO), which focuses on solving two core problems: generation sampled in strong modes of the learned distribution and quality degradation when promoting diversity.
The core idea of TPSO is to jointly operate in two complementary representation spaces.
First, it optimizes learnable offsets in the token embedding space, enabling exploration of underutilized token representations, reducing the tendency to repeatedly sample from strong modes of learned distribution, thereby encouraging diverse generation.
Second, this exploration is regulated by semantic alignment losses in a high-level prompt embedding space, which preserves semantic consistency and prevents the structural distortions commonly observed in prior methods.
Importantly, TPSO is an inference-stage optimization module, making it fully training-free, lightweight, and model-agnostic. This design allows TPSO to seamlessly integrate into existing text-to-image pipelines to enhance generative diversity without compromising image quality. 

Our key contributions are summarized as follows:
\begin{itemize}
\item We introduce a generic, lightweight, training-free, and model-agnostic module for enhancing generative diversity in representative diffusion pipelines.

\item We present TPSO, a dual-space optimization module equipped with diversity and semantic alignment losses.
Guided by these losses, TPSO explores underutilized token-level representations via learnable offsets to mitigate strong-mode sampling in the learned distribution, thereby encouraging diverse generation, while regulating semantic deviation in the prompt embedding space to preserve image fidelity.
A coarse-to-fine scheduling strategy is further introduced to inject diversity at early denoising stages while maintaining high-quality refinement in later steps.

\item Extensive experiments on MS-COCO across three representative diffusion backbones demonstrate that TPSO substantially improves sample diversity (from 1.10 to 4.18) while maintaining image quality with minimal inference-time overhead.
\end{itemize}

\section{Related work}
\label{sec:formatting}

\noindent \textbf{Conditional image generation} has attracted substantial attention from both academia and industry due to its ability to synthesize realistic and aesthetically compelling images. This technology supports a wide range of applications in design, entertainment, advertising, and creative content generation~\cite{rombach2022high,ramesh2022hierarchical,sauer2024adversarial}. To improve controllability and alignment with user intent, conditional generative models guide the generation process using explicit input signals such as text~\cite{rombach2022high, Su_2026}, reference images~\cite{saharia2022palette, lugmayr2022repaint, ho2022cascaded,guo2026instancersrrealworldsuperresolutioninstanceaware}, spatial layouts~\cite{CelebAMask-HQ, zheng2023layoutdiffusion, gao2026diffusion} or combinations~\cite{meng2024mm2latent, zhang2023adding}. Although these approaches improve semantic control and fidelity, most existing efforts focus mainly on image quality and conditional alignment, often overlooking diversity, a fundamental attribute of generative models~\cite{astolfi2024consistency}.

\begin{figure*}[htp]
  \centering
  \includegraphics[width=0.97\textwidth]{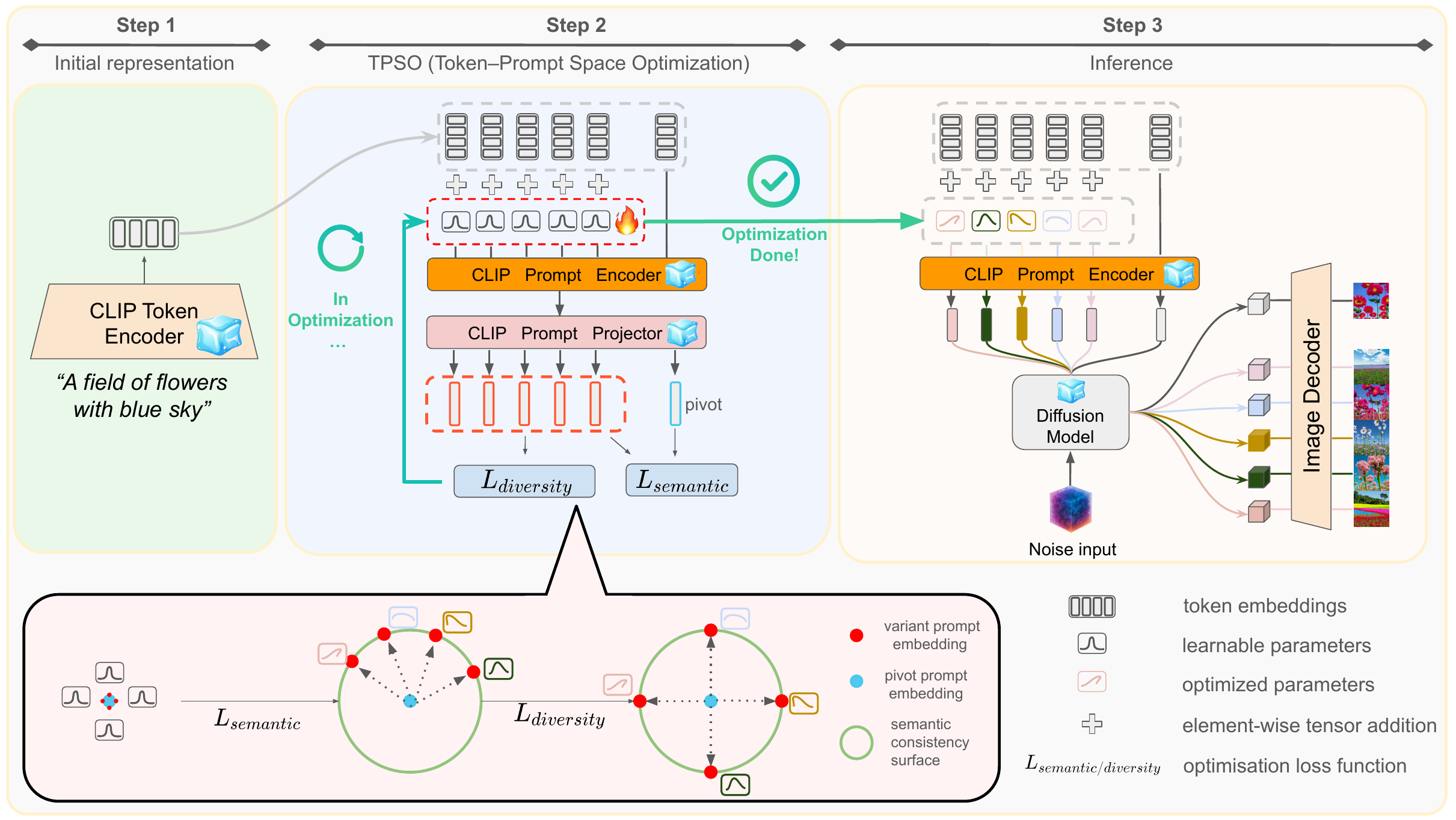}
\caption{
Overview of the TPSO pipeline. TPSO optimizes lightweight learnable offsets in the token embedding space to produce multiple semantically aligned prompt variants, which induce diverse conditional representations and yield different CFG guidance directions. This enables substantially richer generative diversity, all without modifying the diffusion model or introducing any additional components.
}
  \label{fig:pipeline}
\end{figure*}

\noindent \textbf{Diversity in Image generation}
Diversity reflects a generative model’s ability to produce multiple varied outputs under the same conditions and is essential for creative exploration and downstream applications.~\cite{astolfi2024consistency}. However, models often suffer from mode collapse, with samples concentrating around dominant modes~\cite{wu2025indexnet}. Recently, diffusion models~\cite{rombach2022high,song2020denoising} have largely replaced GANs~\cite{karras2019style} as the mainstream generative paradigm, with prior studies~\cite{dhariwal2021diffusion,nichol2021improved} showing that GANs capture less diversity than diffusion models. Existing approaches attempt to enhance diversity in diffusion models by perturbing intermediate latents~\cite{corso2023particle} or adjusting text embeddings~\cite{sadat2023cads}. While effective at separating samples, these techniques often push the generation trajectory into unstable regions of the latent space, resulting in degraded fidelity. A concurrent approach~\cite{trang2025discovering} trains a prompt encoder to generate multiple alternative prompts to increase diversity.
However, it requires additional training and is not explicitly designed to address strong-mode sampling in the learned distribution. In contrast, our TPSO module is entirely training-free and directly explores underutilized token-embedding subspaces to mitigate strong-mode sampling in the learned distribution, while enforcing semantic constraints in the prompt-embedding space to preserve fidelity.
As a result, TPSO achieves increased sample diversity without sacrificing image quality.

\noindent \textbf{Prompt Feature Optimization for Diffusion Models}
\label{param:prompt_optim}
Textual Inversion~\cite{gal2022textual} and MCPL~\cite{jin2024image} optimize new token embeddings in the Stable Diffusion text encoder to represent novel visual concepts.
Similarly, MinorityPrompt~\cite{um2025minority} optimizes token embeddings to improve the generation of minority samples, while ToMe~\cite{hu2024token} merges prompt tokens and optimizes prompt embeddings to address attribute blindness in image generation.
These works indicate that the token embedding space encodes rich semantic information and can be effectively manipulated to influence generative behavior.
Inspired by these approaches, our work introduces token embedding space $\mathcal{S}_{\mathbf{e}} := \{e_i\}_{i=1}^N$ as a new direction for promoting diversity in image generation.
Although CADS~\cite{sadat2023cads} also perturbs representations within the prompt encoder, it operates on the contextualized embeddings $\mathcal{S}_{y} := \{\mathbf{y}_i\}_{i=1}^N$.
In contrast, our method applies optimization directly at the token embedding level $\mathcal{S}_{\mathbf{e}}$, which is more amenable to optimization by leveraging the CLIP text encoder $E$ and the image--text aligned prompt space $\mathcal{S}_{\mathbf{v}} := \{\mathbf{v}_i\}_{i=1}^N$ during optimization.
See \Cref{sec:close_related_work,sec:clip_txt_encoder} for details.

\begin{figure*}[htp]
  \centering
  \includegraphics[width=0.95\textwidth]{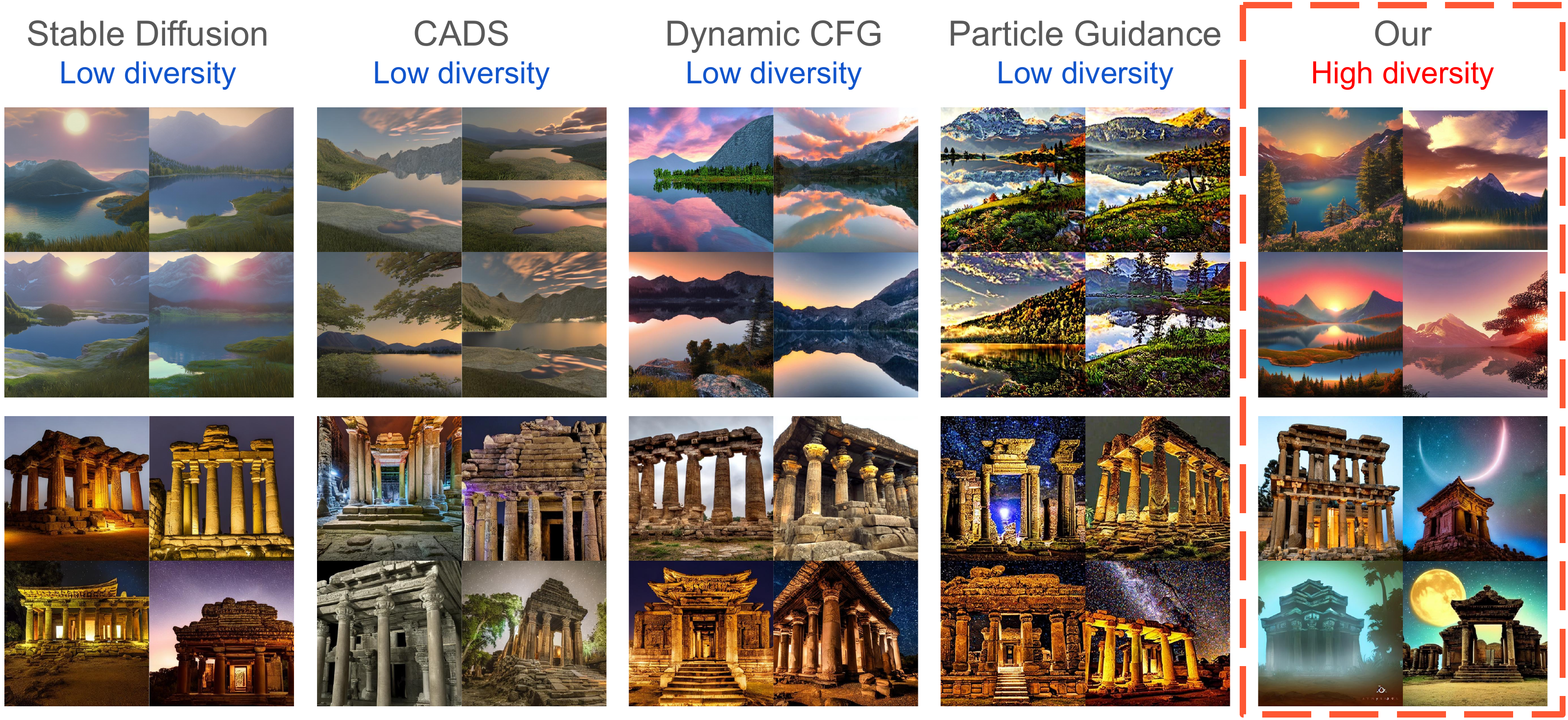}
\caption{TPSO produces a wide range of visually distinct yet semantically faithful variants, demonstrating its ability to increase diversity without sacrificing image quality. In contrast, baseline diffusion models generate highly repetitive outputs across different sampling runs, revealing limited diversity.}

  \label{fig:compare}
\end{figure*}

\section{Preliminary}

\subsection{CLIP Text Encoder}
\label{sec:clip_txt_encoder}
The CLIP text encoder~\cite{Radford2021LearningTV} is a core component in modern text-to-image generation systems, including Stable Diffusion~\cite{rombach2022high,sauer2024adversarial} and DALL·E~\cite{ramesh2022hierarchical}.
Given a text prompt $c$ with tokens $\{i_1,\dots,i_N\}$, each token is first mapped to an embedding via a deterministic lookup,
$\mathbf{e}_n = M[i_n]$, where $M \in \mathbb{R}^{V \times d}$ is the token embedding matrix.
We refer to this stage as the \emph{token encoder}.
The token embeddings are then contextualized by a Transformer-based \emph{prompt encoder} $E$,
yielding $\{y_1,\dots,y_n\} = E(\mathbf{e}_1,\dots,\mathbf{e}_n)$, which serve as the textual conditioning for the diffusion model.
Finally, a lightweight \emph{prompt projector} $f$ maps the hidden state corresponding to the \texttt{<EOS>} token, denoted as $y_{\text{eos}}$, into a high-level image--text aligned semantic space,
$\mathbf{v} = f(y_{\text{eos}})$, this semantic space serves as the target space for our optimization.

\subsection{Quality Degradation in Diversity-Targeted Methods}
\label{sec:close_related_work}
Modern diffusion models often suffer from limited sample diversity~\cite{astolfi2024consistency}.
To mitigate this issue, Particle Guidance~\cite{corso2023particle} and CADS~\cite{sadat2023cads} introduce diversity-promoting perturbations in the latent space and conditioning space of diffusion, respectively.
However, Particle Guidance operates primarily in the latent space at early diffusion steps, where the signal-to-noise ratio is low and semantic information is weak.
Perturbations introduced at this stage are therefore prone to drifting toward low-semantic regions, resulting in degraded image fidelity (see Fig.~\ref{fig:quality_decay}).
CADS promotes diversity by injecting random noise into the diffusion conditioning; although effective, such noise can easily corrupt the conditioning signal and degrade generation quality (Fig.~\ref{fig:quality_decay}).
Consequently, both methods lack an explicit mechanism to preserve fidelity while encouraging diversity.
To address this limitation, we introduce a fidelity-aware diversity module.
Unlike CADS, which random perturbs the contextualized embeddings $y_i$, our TPSO optimizes a \emph{fidelity- and diversity-aware offset} using semantic and diversity losses, which is added at the token embedding level $\mathbf{e}_i$ prior to contextualization (see $\{y_1,\dots,y_n\} = E(\mathbf{e}_1,\dots,\mathbf{e}_n)$, where $\mathbf{e}_i$ denotes the token embedding and $y_i$ its contextualized representation).
This design allows gradients from the projected text representation $\mathbf{v} = f(y_{\text{eos}})$ to propagate through both the prompt encoder $E$ and the projector $f$, enabling effective optimization.
In contrast, perturbations applied directly to $y_i$ in CADS do not affect $y_{\text{eos}}$ and therefore cannot be optimized via objectives defined on $\mathbf{v}$.

\begin{table*}[t]
\centering
\caption{
Main comparison results based on 50k samples. Primary evaluation metrics are highlighted in \textbf{bold}.
}

\label{table:main_compare}

\resizebox{\linewidth}{!}{
\begin{tabular}{lccccccc}
\toprule
& \multicolumn{3}{c}{Quality} & \multicolumn{1}{c}{Consistency} & \multicolumn{3}{c}{\textbf{Diversity}}\\
\cmidrule(lr){2-4} \cmidrule(lr){5-5} \cmidrule(lr){6-8}
Method & \textbf{FID} $\downarrow$ & Precision $\uparrow$& PickScore $\uparrow$ & CLIP $\uparrow$ & \textbf{Recall} $\uparrow$ & \textbf{MSS} $\downarrow$ & \textbf{Vendi} $\uparrow$\\
\midrule

Stable Diffusion 1.5~\cite{rombach2022high} &20.093 & \textbf{0.610} & \textbf{21.48}& \textbf{31.700} & 0.145 & 0.231 & 5.558 \\

Dynamic CFG~\cite{sadat2023cads} &20.405 & 0.606& 21.04& 31.170 & 0.090 & 0.359 & 2.560 \\

CADS~\cite{sadat2023cads} &19.750 & 0.602& 21.09& 30.940 & 0.313 & 0.194 & 5.933 \\
Particle Guidance~\cite{corso2023particle} &19.990 & 0.572 &21.34& 31.320 & 0.218 & 0.194 & 6.387 \\

\rowcolor{gray!15} TPSO 
& \textbf{18.191} \scriptsize(-1.902)
& 0.608 \scriptsize(-0.002)
&21.18 \scriptsize(-0.3)
& 31.100 \scriptsize(-0.6)
& \textbf{0.374} \scriptsize(+0.229)
& \textbf{0.158} \scriptsize(-0.073)
& \textbf{6.705} \scriptsize(+1.147) \\
          
\midrule

Stable Diffusion 2.1~\cite{rombach2022high} &19.871 & 0.599 &\textbf{21.74}& \textbf{31.880} & 0.117 & 0.247 & 5.237 \\
Dynamic CFG~\cite{sadat2023cads} &24.823  & \textbf{0.601} &21.12& 30.960 & 0.033 & 0.355 & 2.608 \\
CADS~\cite{sadat2023cads} &19.891  & 0.599& 21.22& 31.370 & 0.176 & 0.221 & 5.524 \\
Particle Guidance~\cite{corso2023particle} &19.757  & 0.595&21.73 & 31.840 & 0.125 & 0.243 & 5.298 \\
\rowcolor{gray!15} TPSO 
& \textbf{17.569} \scriptsize(-2.302)
& 0.571 \scriptsize(-0.028)
& 21.34 \scriptsize(-0.4)
& 31.100 \scriptsize(-0.78)
& \textbf{0.430} \scriptsize(+0.313)
& \textbf{0.137} \scriptsize(-0.11)
& \textbf{7.299} \scriptsize(+2.062) \\

\midrule

Stable Diffusion 3.5~\cite{esser2024scaling} &24.567  & \textbf{0.645}& 21.87& 32.070 & {0.006} & 0.369 & 2.828 \\
Dynamic CFG~\cite{sadat2023cads} &23.943  & 0.574 & 21.48& 31.310 & {0.037} & 0.350 & 2.739 \\
CADS~\cite{sadat2023cads} &26.934  & 0.639& 22.17& 31.760 & {0.066} & 0.277 & 4.359 \\
Particle Guidance~\cite{corso2023particle} &24.275  & 0.638 & 22.06& \textbf{32.190} & {0.006} & 0.385 & 2.562 \\
\rowcolor{gray!12} TPSO &\textbf{23.623} \scriptsize(-0.944) & 0.576 \scriptsize(-0.069)& 22.10\scriptsize(-0.77) & 31.930 \scriptsize(-0.14) & \textbf{{0.287}} \scriptsize(+0.281) & \textbf{0.156} \scriptsize(-0.213) & \textbf{7.010} \scriptsize(+4.182) \\

\bottomrule
\end{tabular}
} 

\end{table*}

\section{Method}
Motivated by the effectiveness of token-level representations in influencing generative behavior (\Cref{param:prompt_optim}), we leverage the token embedding space $\mathcal{S}_{\mathbf{e}}$ to promote diversity in image generation.
Specifically, we optimize multiple learnable offsets on the deterministic token embeddings.
Each offset is added to the original token embedding to produce a perturbed embedding, which is then processed by the CLIP prompt encoder to form a variant. These variants are fed into the generative model, resulting in diverse outputs under the same condition.
To ensure that these variants maintain both diversity and fidelity, the offsets are jointly optimized using a diversity loss and a semantic alignment loss.
Details of the token and prompt representations are given in \Cref{sec:token_prompt}, followed by the semantic regularization (\Cref{sec:semantic_reg}), the diversity objective (\Cref{sec:diversity_obj}), and the overall optimization and scheduling strategy across diffusion timesteps (\Cref{sec:converage,sec:scheduler}).

\begin{algorithm}[t]
  \caption{Optimization of Learnable Token Offsets}
  \label{algo:optimization}
  \begin{algorithmic}[1]
    \STATE \textbf{Input:} Token embeddings $\boldsymbol{e}_{\text{token}}$, learnable offsets $\boldsymbol{\varepsilon}$
    \STATE \textbf{Output:} Optimized learnable offsets $\boldsymbol{\varepsilon}$
    \STATE Initialize $\boldsymbol{\varepsilon} \sim \mathcal{N}(0, 10^{-4})$

    \STATE Compute $\mathbf{v} = f(E(\boldsymbol{e}_{\text{token}}))$
    \STATE Compute $\mathbf{v}' = f(E(\boldsymbol{e}_{\text{token}} + \boldsymbol{\varepsilon}))$ \hfill (Eq.~\ref{eq:f_ei} and~\ref{eq:semantic_embeddings})

    \REPEAT
        \STATE Update $\boldsymbol{\varepsilon} \gets \text{Adam}(\boldsymbol{\varepsilon}, \nabla_{\varepsilon} \mathcal{L}_{\text{joint}})$ \hfill (Eq.~\ref{eq:loss_joint})
        \STATE Recompute $\mathbf{v}' = f(E(\boldsymbol{e}_{\text{token}} + \boldsymbol{\varepsilon}))$
    \UNTIL $\mathcal{L}_{\text{semantic}}$ convergence (Eq.~\ref{eq:converage})

    \STATE \textbf{return} $\boldsymbol{\varepsilon}$
  \end{algorithmic}
\end{algorithm}

\subsection{Variant Token and Prompt Embedding}
\label{sec:token_prompt}

As illustrated in~\Cref{fig:pipeline}, our method operates on the text encoder of Stable Diffusion. Let $\{i_1, \dots, i_n\}$ denote the input prompt tokens, and let
\begin{equation}
\boldsymbol{e_{\text{token}}} = \{\mathbf{e}_1, \dots, \mathbf{e}_n\}, \qquad \mathbf{e}_i \in \mathbb{R}^d,
\end{equation}
be their corresponding fixed \textsc{CLIP} token embeddings obtained via deterministic matrix indexing $\mathbf{e}_n = M[i_n]$, where $M \in \mathbb{R}^{V \times d}$ (~\Cref{sec:clip_txt_encoder}).

To enable controllable variation, we introduce a set of learnable token-level offset parameters
\begin{equation}
\label{eq:epsilon_def}
\boldsymbol{\varepsilon} = \{\varepsilon_1, \dots, \varepsilon_n\}, 
\qquad 
\varepsilon_i \in \mathbb{R}^d.
\end{equation}

and define the resulting variant token embeddings as
\begin{equation}
\label{eq:f_ei}
\boldsymbol{e}_{\text{token}} + \boldsymbol{\varepsilon}
=
\{\mathbf{e}_1 + \varepsilon_1,\;
  \mathbf{e}_2 + \varepsilon_2,\;
  \dots,\;
  \mathbf{e}_n + \varepsilon_n\}.
\end{equation}

These variant embeddings are then passed through the CLIP prompt encoder $E$ and projector network $f$ to obtain prompt-level embeddings:
\begin{equation}
\label{eq:semantic_embeddings}
\mathbf{v} = f(E(\boldsymbol{e}_{\text{token}})), 
\qquad 
\mathbf{v}' = f(E(
\boldsymbol{e}_{\text{token}} + \boldsymbol{\varepsilon})),
\end{equation}
where both $\mathbf{v},\mathbf{v}' \in \mathbb{R}^{d}$ represent the original and offset-induced prompt embeddings, respectively. These embeddings constitute the basis for the semantic alignment and diversity objective.

\subsection{Semantic Optimization Target}
\label{sec:semantic_reg}
To ensure that the generated variants preserve the semantic meaning of the original prompt, we define a semantic alignment loss.
Since the CLIP prompt embedding space is explicitly structured by cosine similarity (e.g., CLIP Score~\cite{Radford2021LearningTV}), we measure semantic deviation using the cosine similarity between the original prompt embedding $\mathbf{v}$ and each offset-induced embedding $\mathbf{v}_i'$:

\begin{equation}
\label{eq:loss_semantic}
\mathcal{L}_{\text{semantic}}
=
\sum_{i=1}^{N} \big| \cos(\mathbf{v}, \mathbf{v}_i') - \kappa \big| .
\end{equation}
We use a summation rather than an average since each variant has independent learnable offsets, and the gradient from $\mathcal{L}_{\text{semantic}}$ is not shared across variants.

\subsection{Diversity Optimization Target}
\label{sec:diversity_obj}

To encourage the generated image variants to be diverse rather than repeat, we introduce a diversity loss applied to the set of variant prompt embeddings \( \mathbf{v}' = \{\mathbf{v}_1', \dots, \mathbf{v}_N'\} \). This loss penalizes high similarity between any pair of variants:

\begin{equation}
\label{eq:loss_diversity}
\mathcal{L}_{\text{div}}
=
\frac{1}{N(N - 1)}
\sum_{i=1}^{N}
\sum_{\substack{j=1 \\ j \ne i}}^{N}
\cos(\mathbf{v}_i', \mathbf{v}_j').
\end{equation}

\subsection{Convergence Condition}
\label{sec:converage}

The overall objective combines semantic preservation and diversity enhancement:
\begin{equation}
\label{eq:loss_joint}
\mathcal{L}_{\text{joint}}
=
\mathcal{L}_{\text{semantic}}
+
\lambda \cdot \mathcal{L}_{\text{div}},
\end{equation}
where $\lambda$ controls the strength of the diversity term.
The complete optimization procedure is summarized in Algorithm~\ref{algo:optimization}.

For each learnable offset, the optimization terminates when either the convergence criterion in Eq~\ref{eq:converage} is satisfied with tolerance $\sigma = 10^{-2}$, or a maximum of 50 iterations is reached:
\begin{equation}
\label{eq:converage}
\big| \cos(\mathbf{v}, \mathbf{v}_i') - \kappa \big| < \sigma .
\end{equation}

\subsection{Progressive Embedding Scheduler.}
\label{sec:scheduler}
Diffusion sampling follows a coarse-to-fine process~\cite{decatur2025reusing}: early steps determine the global structure and benefit most from optimized embeddings to promote diversity, whereas later steps refine texture and appearance and are best guided by the original embedding to preserve visual quality~\cite{sadat2023cads}.
We exploit this property using a simple ratio-based scheduler.
Diffusion sampling proceeds from timestep $T$ (e.g., $1000$) down to $1$.
Given a proportion parameter $r \in [0,1]$, the early timesteps (from $t = T$ down to $T(1-r)$) use the optimized embedding with a progressive interpolation back to the original embedding, while the remaining timesteps rely solely on the original embedding.
Let $y$ and $y'$ denote the original and optimized embeddings, respectively:

\begin{equation}
y_t^{*} = \alpha_t\, y' + (1 - \alpha_t)\, y,
\end{equation}

where

\begin{equation}
\alpha_t =
\begin{cases}
0, & t \le T(1-r),\\[6pt]
\dfrac{t - T(1-r)}{rT}, & T(1-r) < t \le T .
\end{cases}
\label{eq:schedule}
\end{equation}

We find that setting \( r = 0.4\) provides high diversity while preserving image quality.

\begin{table}[t]
\scriptsize
\centering
\caption{Impact of Semantic Retention Degree $\kappa$}
\label{tab:kappa}

\resizebox{\linewidth}{!}{%
\begin{tabular}{lcccccc}
\toprule
$\kappa$ & FID~$\downarrow$& Precision~$\uparrow$ & CLIP~$\uparrow$ & Recall~$\uparrow$ & MSS~$\downarrow$ & Vendi~$\uparrow$\\
\midrule
0.70 &49.69 & 0.59 & 30.61 & \textbf{0.48} & \textbf{0.12} & \textbf{7.56} \\
0.80 &\textbf{49.58} & 0.60 & 31.14 &  0.37 & 0.16 & 6.65 \\
0.90 &52.30 & \textbf{0.62} & \textbf{31.56} & 0.22 & 0.23 & 5.21 \\

\bottomrule
\end{tabular}

}
\end{table}

\begin{figure}[htp]
  \centering
  \includegraphics[width=0.42\textwidth]{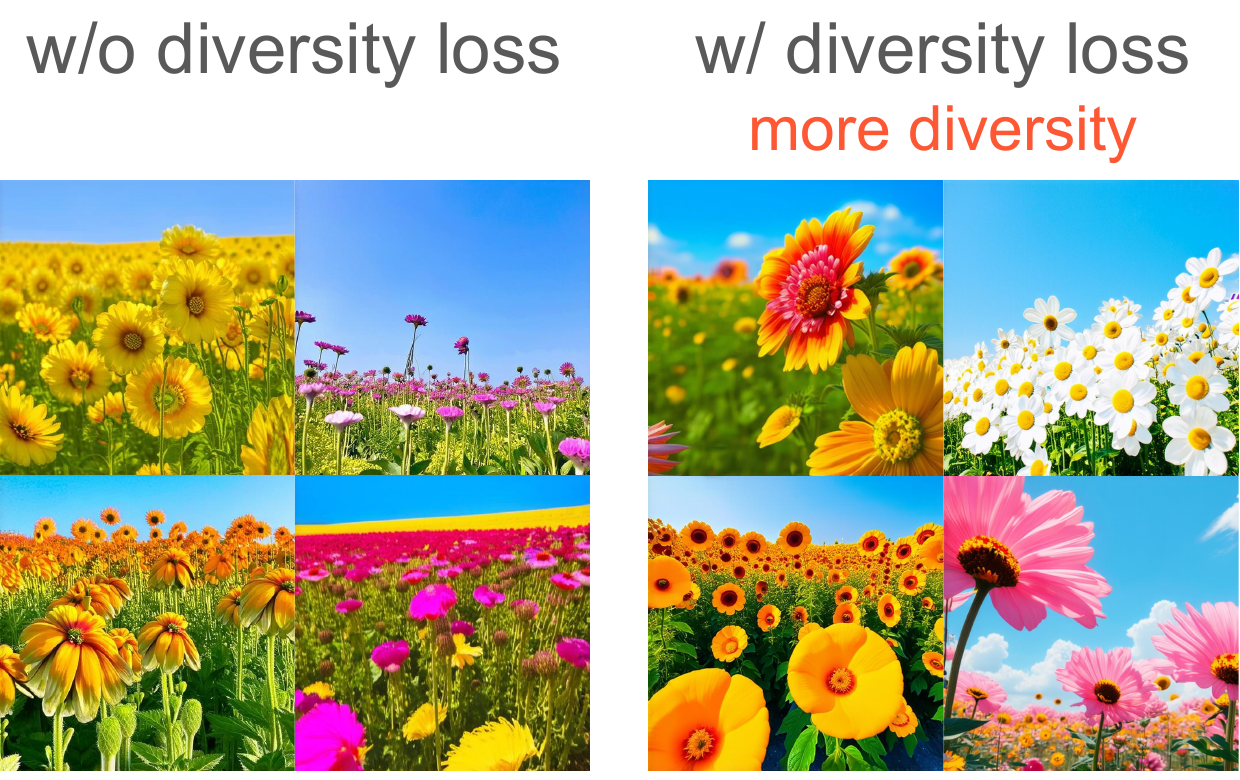}
\caption{Effect of the diversity loss on generation diversity.}

  \label{fig:diversity_loss}
\end{figure}

\section{Experiments}
\label{sec:exp}

\subsection{Experimental Settings}

\noindent \textbf{Implementation Details.}
To verify that TPSO is model-agnostic, we evaluate it on three representative text-to-image generators:
(1) Stable Diffusion 1.5 (SD1.5) and Stable Diffusion 2.1 (SD2.1), both based on the denoising diffusion framework~\cite{rombach2022high,ho2020denoising,song2020denoising}; and
(2) Stable Diffusion 3.5 (SD3.5)~\cite{esser2024scaling}, which adopts a rectified-flow formulation~\cite{liu2022flow,lipman2022flow}.
We follow the default public configurations for all models.
Notably, SD3.5~\cite{esser2024scaling} employs two CLIP text encoders, which we optimize independently.
All experiments are conducted with fixed random seeds to ensure fair and model-consistent comparisons on an A100 GPU using PyTorch 2.6.

\begin{table}[t]
\centering
\caption{Comparison of coarse-to-fine and fine-to-coarse scheduling for optimized embeddings.}

\label{tab:schedule_compare}
\renewcommand{\arraystretch}{1.1}
\begin{tabular}{lcccc}
\toprule
\textbf{Scheduling Strategy} & Setting &
FID~$\downarrow$ & CLIP~$\uparrow$ & Recall~$\uparrow$ \\
\midrule
\multirow{2}{*}{\textbf{Coarse-to-Fine~($r>0$)}} 
& $r = 0.4$ & \textbf{49.58} & 31.14 & 0.374 \\
& $r = 0.6$ & 50.06 & 30.96 & \textbf{0.433} \\
\midrule
\multirow{2}{*}{\textbf{Fine-to-Coarse~($r<0$)}} 
& $r = -0.4$ & 63.53 & \textbf{31.63} & 0.165 \\
& $r = -0.6$ & 61.55 & 31.59 & 0.169 \\
\bottomrule
\end{tabular}
\end{table}

\noindent \textbf{Dataset.}
Following the experimental protocol of CADS~\cite{sadat2023cads}, we conduct all experiments on the MSCOCO validation set~\cite{lin2014microsoft}.
We randomly sample 5,000 captions and generate 10 images per caption, resulting in 50k samples for each model–method pair. For ablation studies, we use a reduced subset of 1,000 captions (10k samples).

\subsection{Evaluation Metrics}
We follow CADS~\cite{sadat2023cads} and evaluate our method across three complementary aspects: \textbf{quality}, \textbf{conditional alignment}, and \textbf{diversity}.

\begin{table}[t]
\centering
\caption{
Effect of different initialization distributions for the learnable parameter $\varepsilon$ (Line~3 of \Cref{algo:optimization}).
}
\fontsize{8}{8}\selectfont
\resizebox{\linewidth}{!}{
\begin{tabular}{lcccccc}
\toprule
& \multicolumn{2}{c}{Quality} & \multicolumn{1}{c}{Consistency} & \multicolumn{3}{c}{Diversity} \\
\cmidrule(lr){2-3} \cmidrule(lr){4-4} \cmidrule(lr){5-7}
Initialization  & FID $\downarrow$ & Precision $\uparrow$ & CLIP $\uparrow$ & Recall $\uparrow$ & MSS $\downarrow$ & Vendi $\uparrow$ \\
\midrule
Normal(0, 1e-4)  & \textbf{50.69} & 0.560 & \textbf{30.06} & 0.508 & 0.121 & 7.76 \\
Normal(0, 1)     & 52.82 & 0.535 & 29.02 & 0.552 & 0.117 & 7.87 \\
\midrule
Laplace(0, 1)          & 59.97 & 0.494 & 27.84 & 0.551 & 0.097 & 8.30 \\
Laplace(0, $\sqrt{1.5}$) & 67.22 & 0.459 & 27.02 & \textbf{0.554} & \textbf{0.087} & \textbf{8.52} \\
\midrule
Gamma(1,1)               & 54.73 & 0.539 & 28.69 &  0.532& 0.109 & 8.05 \\
Gamma(4,$\sqrt{4}$)      & 53.83 & 0.543 & 28.86 & 0.533 & 0.111 & 8.00 \\
\midrule
Uniform(-1,1)   & 51.10 & \textbf{0.563} & 29.83 & 0.51 & 0.125 & 7.70 \\
Uniform(-3,3)   & 63.98 & 0.466 & 27.44 & 0.547 & 0.092 & 8.39 \\
\bottomrule
\end{tabular}
}
\label{table:param_initialization}
\end{table}

\noindent \textbf{Diversity Metrics.}
We report Recall~\cite{kynkaanniemi2019improved} as a set-level diversity metric between synthetic and real images.
Sample-level diversity is measured using the Mean Similarity Score (MSS) and the Vendi Score~\cite{friedman2022vendi}, which assess diversity among samples generated under the same condition. Both metrics are computed with SSCD~\cite{pizzi2022self} features following CADS~\cite{sadat2023cads}.

\noindent \textbf{Quality Metrics.}
Fréchet Inception Distance (FID)~\cite{Heusel2017GANsTB} serves as our primary quality metric.
We also report Precision~\cite{kynkaanniemi2019improved} as a complementary measure of sample fidelity.
However, as noted in CADS, Precision does not reliably evaluate models that produce highly diverse outputs.

\noindent \textbf{Alignment Metric.}  
Text–image consistency is evaluated using the CLIP Score~\cite{Radford2021LearningTV}, defined as the cosine similarity between CLIP embeddings of the generated image and its corresponding prompt.

\subsection{Qualitative Results.}
\noindent \textbf{Comparison with Baseline Methods.}
Our objective is to increase image diversity while preserving visual fidelity.
As shown in Fig.~\ref{fig:compare}, baseline methods tend to produce visually similar results, often repeating comparable layouts or structures even when different noise seeds are used.
In contrast, our method generates a broader range of plausible variations, exhibiting differences in geometry, color, and fine-grained details while remaining faithful to the input prompt.

Importantly, TPSO achieves this diversity while maintaining image fidelity, unlike Dynamic CFG, CADS, and Particle Guidance, which often improve diversity at the expense of text alignment or visual quality.

\begin{table}[t]
\scriptsize
\centering
\caption{Impact of the diversity-loss weight $\lambda$.}
\label{tab:lambda}
\begin{tabular}{lcccccc}
\toprule
$\lambda$ & FID~$\downarrow$& Precision~$\uparrow$ & CLIP~$\uparrow$ & Recall~$\uparrow$ & MSS~$\downarrow$ & Vendi~$\uparrow$\\
\midrule
0.0 &49.58 & 0.60 & \textbf{31.14} &  0.37& 0.16 & 6.65 \\
5.0  &\textbf{49.38}  & \textbf{0.62} & 31.04 & 0.43 & 0.14 & 7.13 \\
10.0 &48.54 & \textbf{0.62} & 30.89 & \textbf{0.47} & \textbf{0.12} & \textbf{7.54} \\

\bottomrule
\end{tabular}
\end{table}

\begin{figure}[t]
    \centering
    \includegraphics[width=0.8\linewidth]{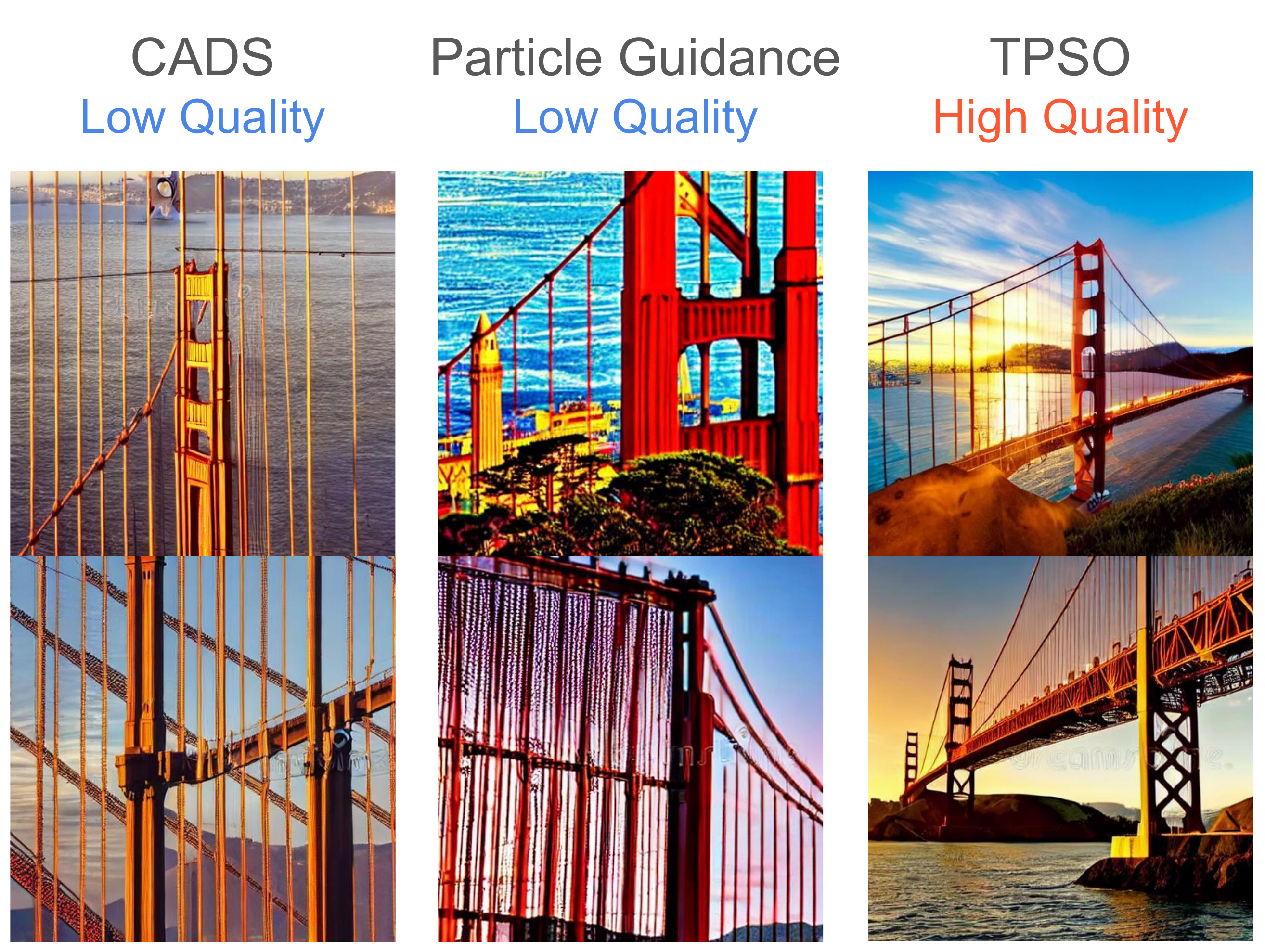}
\caption{
Quality preservation under increased diversity.
TPSO maintains higher visual quality and semantic alignment than CADS and Particle Guidance.
}

\label{fig:quality_decay}
\end{figure}

\noindent \textbf{Quality Preservation}
\label{sec:visual_quality}
As shown in Fig.~\ref{fig:quality_decay}, CADS~\cite{sadat2023cads} and Particle Guidance~\cite{corso2023particle} introduce greater structural distortions or semantic drift than TPSO when increasing diversity, highlighting the challenge of improving diversity without degrading image fidelity.

\setlength{\tabcolsep}{3pt}
\begin{table}[htb]
    \caption{Ablation study of $\mathcal{L}_{\text{semantic}}$ and $\mathcal{L}_{\text{diversity}}$. The first row adds a learnable offset $\boldsymbol{\varepsilon} \sim \mathcal{N}(0, 10^{-4})$ without optimization.}

	\centering
	\begin{tabular}{c c c c c c c}
		\toprule
		Optimization & $\mathcal{L}_{\text{semantic}}$ & $\mathcal{L}_{\text{diversity}}$ & FID $\downarrow$ & CLIP $\uparrow$ & Recall $\uparrow$ & Vendi $\uparrow$ \\
		\midrule
		\xmark & \xmark & \xmark & 58.11 & \textbf{31.61} & 0.03 & 1.98 \\
		\checkmark & \xmark & \checkmark &  55.42&  28.47&  \textbf{0.62}& \textbf{9.21} \\
		\checkmark & \checkmark & \xmark &  49.58&  31.14&  0.37& 6.65   \\
        \rowcolor{gray!15} \checkmark & \checkmark & \checkmark &  \textbf{49.38}&  31.04&  0.43& 7.13  \\
		\bottomrule
	\end{tabular}
	\label{tab:ablation_lsemantic_ldiverse_optimize}
\end{table}
\subsection{Quantitative Results}
\Cref{table:main_compare} summarizes the quantitative performance across all baselines and diffusion backbones.
TPSO consistently improves all diversity metrics (Recall, MSS, and Vendi) while also achieving better FID, our primary quality indicator.
These results demonstrate that TPSO effectively enhances diversity without sacrificing overall image fidelity.

Although Precision exhibits a slight decrease, this reflects a known limitation of the metric, as it can favor high-quality yet less diverse samples~\cite{sadat2023cads}. This should not be interpreted as a degradation in image quality.

TPSO shows minor reductions in CLIP Score. This is because our method introduces learnable offsets in the prompt embedding space, which may induce small but controlled semantic shifts.
However, these changes are marginal (less than 0.78 in absolute difference) and negligible compared to the substantial gains in diversity and FID.

\subsection{Ablation Study}
\noindent \textbf{Semantic Retention Degree $\kappa$.}
$\kappa$ controls the semantic alignment between variant and original prompt embeddings.
Lower $\kappa$ permits larger token-level deviations, resulting in reduced fidelity but increased diversity (\Cref{tab:kappa}).
We find $\kappa=0.80$ provides a strong balance between diversity and semantic fidelity.

\noindent \textbf{Effect of Diversity Loss and Weight $\lambda$.}
As shown in~\Cref{tab:lambda}, larger diversity-loss weights $\lambda$ induce stronger intra-prompt variation, leading to more diverse generations.

\begin{table}[t]
\centering
\caption{Inference time across Stable Diffusion versions.}
\resizebox{\linewidth}{!}{
\begin{tabular}{lccc}
\toprule
& \multicolumn{3}{c}{\textbf{Time (sec / sample)}~$\downarrow$}\\
\cmidrule(lr){2-4}
\textbf{Method} 
& \textbf{SDv1.5}  
& \textbf{SDv2.1} 
& \textbf{SDv3.5} \\
\midrule
Stable Diffusion (SD)~\cite{rombach2022high,esser2024scaling}  
& \textbf{0.78} & \textbf{1.80} & \textbf{7.46} \\
Dynamic CFG~\cite{sadat2023cads} 
& 0.79 & 1.80 & 7.46 \\
CADS~\cite{sadat2023cads}              
& 0.80 & 1.80 & 7.47 \\
Particle Guidance~\cite{corso2023particle} 
& 0.79 & 1.80 & 7.46 \\
\rowcolor{gray!12}
\rowcolor{gray!12}
TPSO 
& 0.85 {\scriptsize(+8.9\%)} 
& 1.93 {\scriptsize(+7.2\%)} 
& 7.73 {\scriptsize(+3.6\%)} \\
\bottomrule
\end{tabular}}
\label{tab:speed_sd1_2_3}
\end{table}
\noindent \textbf{Effect of $\mathcal{L}_{\text{semantic}}$ and $\mathcal{L}_{\text{diversity}}$.}
Tab.~\ref{tab:ablation_lsemantic_ldiverse_optimize} shows that using $\mathcal{L}_{\text{diversity}}$ alone increases diversity at the cost of fidelity, while using $\mathcal{L}_{\text{semantic}}$ alone yields limited diversity.
Joint optimization achieves the best diversity–fidelity trade-off.

\noindent \textbf{Effect of the Coarse-to-Fine Scheduler.}
The ratio $r$ controls the portion of the sampling trajectory where optimized embeddings are applied.
Positive values (e.g., $r=0.4$) apply optimized embeddings in early denoising stages, whereas negative values (e.g., $r=-0.4$) apply them in later refinement stages, serving as a control for the coarse-to-fine design.

As shown in \Cref{tab:schedule_compare}, applying variations at early stages ($r>0$) significantly improves diversity while preserving visual quality, whereas late-stage variations ($r<0$) yield limited diversity gains and noticeably degrade image quality.
These results support the effectiveness of the proposed coarse-to-fine scheduling strategy.

\noindent \textbf{Parameter Initialization.}
Initialization affects optimization stability and convergence~\cite{glorot2010understanding}. We compare different initialization strategies in Tab.~\ref{table:param_initialization}, showing that $\mathcal{N}(0,10^{-4})$ yields the best performance.

\noindent \textbf{Inference Efficiency.}
Our method incurs only a small inference-time overhead of 3.6\%–8.9\% across Stable Diffusion variants (Table~\ref{tab:speed_sd1_2_3}).

\section{Limitations and Future Work}
\noindent \textbf{Limitations.} TPSO remains bounded by the representational capacity of the underlying diffusion model, as it does not modify model parameters. It relies on CLIP embeddings for semantic alignment and is therefore limited by CLIP’s capacity. Large perturbations may still cause mild quality degradation in rare cases. TPSO also introduces a small inference-time overhead due to per-prompt optimization.

\noindent \textbf{Future Work.} Although TPSO operates at inference time, its effectiveness suggests potential extensions to training. Incorporating embedding-space exploration or diversity-aware regularization during training may reduce mode collapse and improve coverage, representing a promising direction for future research.

\section{Conclusion}
\label{sec:conclusion}
We present TPSO, a training-free and model-agnostic module for enhancing diversity in text-to-image generation without sacrificing fidelity.
By optimizing learnable offsets at the token level and regulating semantic deviation in the prompt space, TPSO enables controlled exploration of underrepresented regions of the token embedding space, reducing repeated sampling from strong modes of the distribution.
Diversity is further enhanced through an explicit diversity loss.
A coarse-to-fine scheduling strategy introduces diversity at early denoising stages while preserving high-quality refinement in later steps.
Extensive experiments on SD1.5, SD2.1, and SD3.5 demonstrate consistent improvements in per-prompt and global diversity metrics, while preserving image fidelity with minimal inference-time overhead.
Given its simplicity, generality, and compatibility with modern diffusion and rectified-flow models, TPSO provides a practical solution for diversity-enhanced image generation.

{\footnotesize
\noindent \textbf{Acknowledgements}
DM's and GT's work was funded by UK Research and Innovation (UKRI) under the UK government’s Horizon Europe funding guarantee (grant No. 10099264) and by the European Union (under EC Horizon Europe grant agreement No. 101135800 (RAIDO)). For the purpose of open access, the author has applied a Creative Commons Attribution (CC BY) license to any Author Accepted Manuscript version arising.\par
}

\end{document}